\def\year{2021}\relax
\documentclass[letterpaper]{article} % DO NOT CHANGE THIS
\usepackage{aaai21}  % DO NOT CHANGE THIS
\usepackage{times}  % DO NOT CHANGE THIS
\usepackage{helvet} % DO NOT CHANGE THIS
\usepackage{courier}  % DO NOT CHANGE THIS
\usepackage[hyphens]{url}  % DO NOT CHANGE THIS
\usepackage{graphicx} % DO NOT CHANGE THIS
\usepackage{natbib}  % DO NOT CHANGE THIS AND DO NOT ADD ANY OPTIONS TO IT
\usepackage{caption} % DO NOT CHANGE THIS AND DO NOT ADD ANY OPTIONS TO IT
\usepackage{amsfonts}
\usepackage{amsmath}
\usepackage{amsthm}  
\usepackage{dsfont}  
\usepackage{subcaption}
\usepackage{cleveref}
\usepackage{dsfont} 
\usepackage[ruled]{algorithm2e} % For algorithms

\usepackage{graphicx}
\usepackage{wrapfig}

\newtheorem{thm}{Theorem}

\title{Cascade Size Distributions: Why They Matter and How to Compute Them Efficiently}
\author{
     Rebekka Burkholz, \textsuperscript{\rm 1}
     John Quackenbush \textsuperscript{\rm 1} \textsuperscript{\rm 2}\\
}
\affiliations{
\textsuperscript{\rm 1} Department of Biostatistics, 
  Harvard T.H. Chan School of Public Health, 
  Boston, MA 02115\\
  \textsuperscript{\rm 2} Harvard Medical School, 
  Boston, MA 02115\\
}
\begin{document}

\maketitle

\begin{abstract}
Cascade models are central to understanding, predicting, and controlling epidemic spreading and information propagation. Related optimization, including influence maximization, model parameter inference, or the development of vaccination strategies, relies heavily on sampling from a model. This is either inefficient or inaccurate. As alternative, we present an efficient message passing algorithm that computes the probability distribution of the cascade size for the Independent Cascade Model on weighted directed networks and generalizations. Our approach is exact on trees but can be applied to any network topology. It approximates locally tree-like networks well, scales to large networks, and can lead to surprisingly good performance on more dense networks, as we also exemplify on real world data. 
\end{abstract}
\section{Introduction}

The Independent Cascade Model (ICM) is a cornerstone in the study of spreading processes on networks.  
It has been proven useful in the source detection of epidemic outbreaks \cite{LeskovecKrause,SourceId,rumourICM,multipleSources}, classification of fake news \cite{fakeNews,falseNewsScience}, marketing \cite{viralMarketing,KempeTMandICMequiv}, or identification of causal miRNAs for cancer \cite{cancer}. 
It can be mapped to the SIR (Susceptible Infected Recovered) model \cite{Kermack700}, which together with its variants has served during the COVID-19 pandemic to estimate associated risks in different scenarios \cite{covidScience,Zhangeabb8001}. 
Due to its simplicity, the model is well suitable to estimate a spreading process in situations of high uncertainty. 
It still provides mechanistic insights that allow to predict changes in the spreading dynamics due to different interventions to an ongoing cascade.

The main quantity of interest is the number of infected people, which is also called the cascade size. 
Finding a feasible policy that minimizes the risk of large cascade sizes poses a challenging optimization problem. 
It also requires to estimate model parameters based on highly uncertain data.
Furthermore, the only information available is usually the number of infected people, in particular, at early stages of the disease spreading.
Another common optimization problem related to the ICM is influence maximization, i.e., the maximization of the average cascade size by optimal selection of seeds (i.e., initial spreaders) \cite{Kempe2003}.
These exemplary optimization problems have in common that they require sampling from the probability distribution of the cascade size or variational inference approaches. 
Yet, sampling is computationally costly or inaccurate. 
Variational inference usually relies on Belief Propagation \cite{DMPpre,DBLP:conf/nips/Lokhov16}, which estimates marginals in graphical models, i.e., the infection probability of each node in a network and their average. 
%Hence, it also estimates the average cascade size. 
Yet, it cannot capture the strong positive dependence between node activations that is due to mutual infections. 
For similar reasons, the average cascade size can be a bad representation of the probability distribution, which is often broad and multi-modal \cite{BurkholzFinite,TDAcascades}, see also Fig.~\ref{aig:NumTheo}.

We propose an accurate and efficient alternative that is based on message passing like Belief Propagation but follows different principles, which we term Subtree Distribution Propagation (SDP).
Subtree Distribution Propagation generalizes to discrete models with more than two states and can capture many SIR-type spreading processes.  
In contrast to Belief Propagation, we provide information on joint activations by computing the full probability distribution of the final cascade size (instead of the average cascade size).
In addition, we obtain the infection probabilities conditional on the cascade size (instead of the unconditional ones). 

Up to our knowledge, conditional infection probabilities have not been studied before but provide rich information about the spreading behavior and susceptibility of nodes.  
1) They allow to identify nodes that are functionally similar with respect to the cascade process.
These similarities are usually caused by network symmetries.
In a biological setting, these can imply redundant pathways and are particularly relevant for deepening our understanding of diseases like cancer \cite{otter}. 
2) Hence, they define a natural node embedding where the embedding space corresponds to the final cascade size distribution. 
Note that it is common practice to compare node embeddings by using them to estimate the infection probabilities of nodes in SIR-type spreading processes.
As we show, these probabilities can also be used to define a node embedding directly. 
3) Identifying node similarities (for instance by node clustering) has also algorithmic implications. 
They reduce the set of seed candidates in influence maximization or related problems and thus speed up Greedy approaches.   
In some cases, they can even enable exhaustive search (see supplementary material).  

As Belief Propagation, the proposed algorithms are exact on trees.
They require maximally $O(N^2)$ computations, where $N$ denotes the number of network nodes, but the run time scales usually as $O(N)$ on sparse networks. 
Parallelization can speed this up further. %speed-up computation. 
As extension to general networks, we propose Tree Distribution Approximation (TDA).
It combines Belief Propagation (BP) with Subtree Distribution Propagation (SDP). 
TDA is approximate, but accurate on locally tree-like networks (like BP). 
Beyond performance gains, our algorithms have the advantage over sampling that they provide us with functional relationships between cascade model parameters and outputs, i.e. the cascade size distribution and conditional infection probabilities.
These would allow us to also compute gradients with respect to model parameters  efficiently and thus enable first-order (instead of zeroth-order) optimization approaches. 

\subsection{The Independent Cascade Model}
The ICM models the binary, stochastic, and discrete activation dynamics of nodes in an undirected network $G = (V, E)$ consisting of $N = |V|$ nodes that are connected by links in $E$.
Each node $i$ is either inactive $(s_i = 0)$ or active $(s_i = 1)$ and can only switch from an inactive to an active state, but not vice versa.  
Initially, each node $i$ activates with probability $p_i$ independently of the other nodes. 
In the next time step $(t=1)$, an active node $i$ can trigger new activations of its neighbors $j \in \rm{nb}(i) := \{j \mid (i,j) \in E\}$. 
Its degree $d_i = |\rm{nb}(i)|$ counts the number of neighbors it has.   
Each neighbor $j$ activates independently with probability $w_{ij}$ and can cause new activations in the next time step. 
This way, a cascade keeps propagating, where several activations can happen at the same time $t$ and each node becoming active at $t$ can trigger new activations only in the next time step $t+1$ but not any later times.
The process ends at time $T \leq N$, when no further activations can occur. 
Then, the fraction of active nodes $\rho = 1/N \sum^N_{i=1} s_i(T)$ defines the final cascade size. 
This is the realization of a random variable $C$ with probability distribution $p_C(\rho)$, as the cascade process is stochastic.
$p_C(\rho)$ is also termed probability mass function of $C$ and has support $\{0, 1/N, ..., 1\}$. 
In summary, an ICM is parametrized by $(\mathbf{p}, \mathbf{W})$, where the vector $\mathbf{p}$ has components $p_i$ and the matrix $\mathbf{W}$ entries $w_{ij}$.
Note that $W$ is often related to network weights and can encode directedness by $w_{ij} \neq w_{ji}$ and $w_{ij} = 0$. 

\subsection{Contributions}

Our main contribution is the development of efficient message passing algorithms that compute the final cascade size distribution and node infection probabilities conditional on the final cascade size for the Independent Cascade Model. 
%that provide probability distribution information 
%for the computation of the final cascade size distribution for the Independent Cascade Model. 
Subtree Distribution Propagation (SDP) is exact on trees and requires $O(N^2)$ computations.
Tree Distribution Propagation (TDA) applies to general networks and leads to accurate inference of the cascade size distribution on locally tree-like networks.
Additional backpropagation computes the infection probabilities of nodes conditional on the final cascade size. 
As we show in experiments, our algorithms scale favorably with increasing network size and lead to speed-ups by a factor ranging from $60-500$ in comparison with sampling. 
Efficiency gains are not only relevant for modeling spreading phenomena on large networks but also on smaller ones if the cascade size distribution is required repeatedly for different parameters, as it is common in most related optimization approaches. 
We further investigate the limitations of our algorithms by studying them on random graphs that are not locally tree-like but find surprisingly good performance. 
In addition, we provide examples on real world data sets, including a dense network of miRNA signaling corresponding to gastrointestinal cancer and a large YouTube network.

\subsection{Related Literature}
As diverse as spreading phenomena are the related optimization objectives. 
The insight that the average cascade size is a submodular influence function \cite{Kempe2003} has inspired many efforts to maximize this quantity by nearly optimal seed size selection \cite{pmlr-v32-du14,resourcesBP} or network adjustments \cite{NipsIM}. 
This has great applications, e.g., in marketing \cite{Morris2000Contagion,Goldenberg2001,Domingos:2001:MNV:502512.502525}. 
%An alternative goal can be to balance information to avoid filter bubbles \cite{infoBalance} or overexposure \cite{mitigatingOverexposures}. 
Other works are more concerned with destructive aspects of cascades and their mitigation to avoid epidemic spreading \cite{limitSpreadMisinformation,resourcesBP}. 
Then, the not always explicit objective is to minimize cascades or to create boundary conditions that limit their size \cite{multiplex,correlations,foodtrade,LRD}.
In some cases, the objective can also be to keep cascades within a specified range \cite{DIP,controlSandpile}.  
Another example for a related optimization problem are maximum likelihood approaches to infer cascade model parameters in the context of information propagation in social networks. %, which is often based on maximum likelihood approaches in the context of information propagation in social networks.  
Several works assume the knowledge of full \cite{NIPS2010_4113,10.5555/3104482.3104553,NIPS2012_4582} or at least partial \cite{DBLP:conf/nips/Lokhov16,10.1145/2487575.2487664} diffusion information, i.e., data on which node becomes infected when, while the network is mostly known or has to be learned \cite{Hoffmann2019,DBLP:journals/corr/abs-1301-6916}. 
Likelihoods are analytically attainable when the time evolution is observed. 
When this information is missing but node identities are known, network recovery is sometimes still feasible \cite{10.5555/3044805.3045098}. 
%Missing time information has also been addressed for network recovery problems \cite{10.5555/3044805.3045098}.
%Still, the node identity is assumed to be known. 
However, this kind of data is usually not available during an epidemic outbreak.
Hence, epidemic spreading models are mainly calibrated by parameter grid search and model sampling \cite{covidScience,Zhangeabb8001} and most of the related optimization approaches are based on average cascade sizes and/or sampling from the cascade model. 

To compute the average cascade size, faster alternatives to sampling are provided by Local Tree Approximations \cite{NewmanSIR} for large random networks or Belief Propagation for smaller sparse networks \cite{BPthresholds,resourcesBP,TDAcascades}. 
For specific optimization problems like influence maximization, also more efficient sampling techniques have been developed, which reuse samples or terminate non-promising simulations in smart ways.
Prominent examples are given by RIS \cite{RIS} and SSA \cite{SSA} sampling. 

Yet, as recently shown \cite{predictCascades,BurkholzFinite,TDAcascades}, average cascade sizes do not provide relevant summary statistics in cases when the underlying cascade size distributions are broad, multi-modal, and support extreme events. 
%In this case, the average does not provide a relevant summary statistic. 
Examples are provided in Fig.~\ref{aig:NumTheo}. 
In every case, the full probability distribution provides much richer information about a network structure. 
It is also necessary for defining the likelihood of observed cascades and related model estimation.
In addition, it allows to estimate the risk of extreme events. 
Usually, those events are of highest interest to judge the robustness of a system or to find optima \cite{Battiston10031,Embrechts,portfolio}. 
For instance, \cite{portfolio} maximizes the expected shortfall of the final cascade size with respect to a portfolio of seeds by sampling from an ICM.  
Often, very large or small cascades occur only with small probabilities.
This makes the optimization of tails harder, in particular, by sampling and demands alternatives. 
For a given simple ICM with uniform infection probability $p_i = p$ or threshold model and a locally tree-like network, the final cascade size distribution can be computed by message passing \cite{TDAcascades} in O($N^2 \log(N)$). 
Our approach is inspired, even though the math is different and our algorithms are more efficient (O($N^2$)).
Furthermore, we can capture general ICMs with heterogeneous weights and initial activation probabilities, while \cite{TDAcascades} is restricted to the uniform case $p_i = w_{ij} = p$. 
In addition, we provide activation probabilities of nodes conditional on the final cascade size.

\section{Motivation: Why cascade size distributions?}
\begin{figure}[t]
\includegraphics[width=0.45\textwidth]{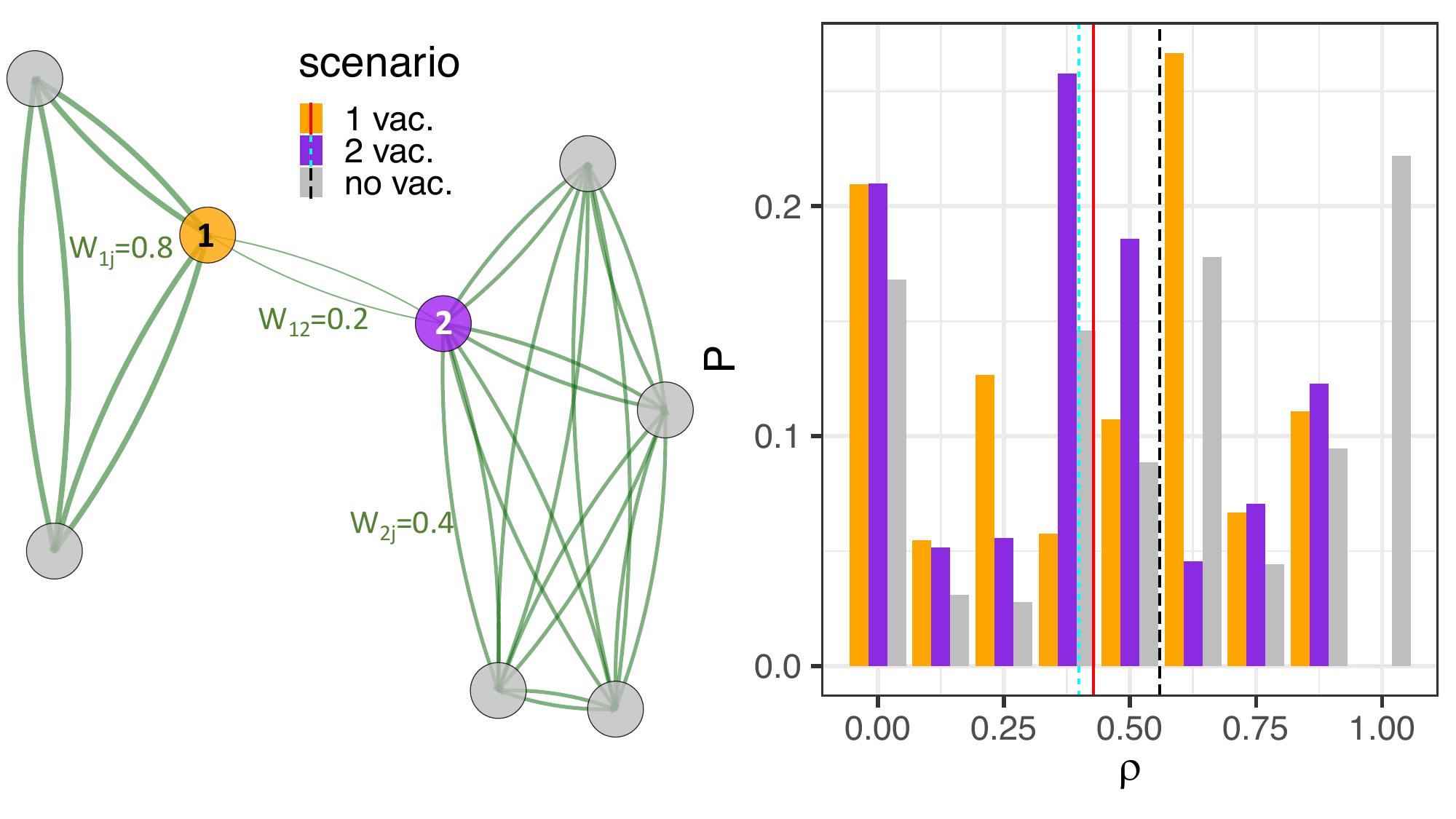}
 \caption{Comparing the average cascade size and tail risk. The histograms show the cascade size distribution in different scenarios for an ICM with $p_i = 0.2$, $w_{12}=w_{21}=0.2$, $w_{ij}=0.8$ within the clique of size 3, and $w_{ij}=0.4$ within the clique of size 5. Lines mark average cascade sizes. Vaccinating Node $2$ minimizes the average cascade size, while vaccinating Node $1$ minimizes the probability of large cascades $\mathbb{P}\left(\varrho \geq 0.75\right)$.
 }\label{fig:toy}
\end{figure}

%To motivate our efforts to compute cascade size distributions efficiently, 
To motivate the benefits of taking cascade size distributions into account, let us consider a toy example of a social network, as depicted in Fig.~\ref{fig:toy}, which is exposed to an epidemic that evolves according to an ICM. 
Without any interventions, the cascade size is distributed as visualized by the gray histogram. The black dashed line marks the average cascade size, which does not correspond to the most probable events and does not prepare policy makers for the high risk of large cascades. 

Next, we assume that we can mitigate the risk with vaccinations.  
Our goal is to avoid large cascades that would exceed the available hospital capacities and to minimize the number of deaths.
Who should be vaccinated first? 
Nodes $1$ and $2$ are the best candidates with identical exposure $\sum_j w_{ij}$.
Node $1$ interacts with less people but has closer contact with them than Node $2$. 
Interestingly, vaccinating $1$, the node with the smaller degree, would minimize the risk of large cascades $\mathbb{P}\left(\varrho \geq 0.75\right)$, while vaccinating $2$ would minimize the average cascade.
If hospital capacities are limited, vaccinating $1$ might be the most promising strategy, which requires the estimation of the cascade size distribution's tail.

\section{Algorithmic approach}
Next, we describe the novel message passing algorithms that we propose to estimate the final cascade size distribution and conditional activation probabilities of nodes.
While we focus in our derivations on the ICM, the principle is in fact more general and could be transferred to continuous time \cite{continuous} and different binary graphical models or more complicated epidemic spreading processes, in which nodes are equipped with a higher number of discrete states.  

\begin{figure}[t]
\includegraphics[width=0.48\textwidth]{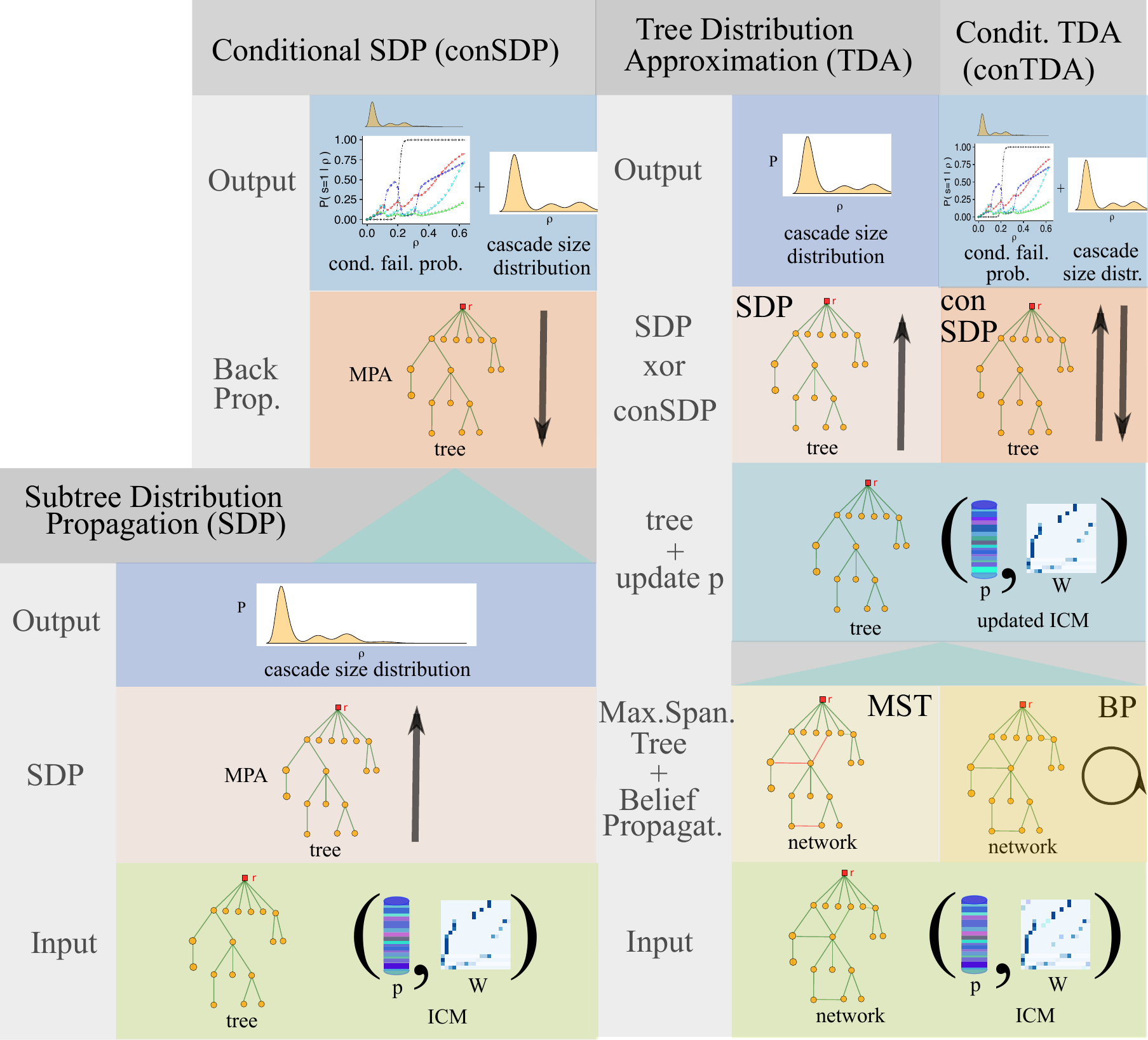}
 \caption{Overview of different message passing algorithm variants for the final cascade size distribution. All are exact on trees, where SDP coincides with TDA and ConSDP with ConTDA. TDA and conTDA are approximative for networks that are different from trees. In addition, conSDP or conTDA variants compute conditional activation probabilities of nodes.  
 }\label{aig:algos}
\end{figure} 

\subsection{Message passing algorithms}
\begin{figure*}[t]
\begin{center}
\includegraphics[width=0.75\textwidth]{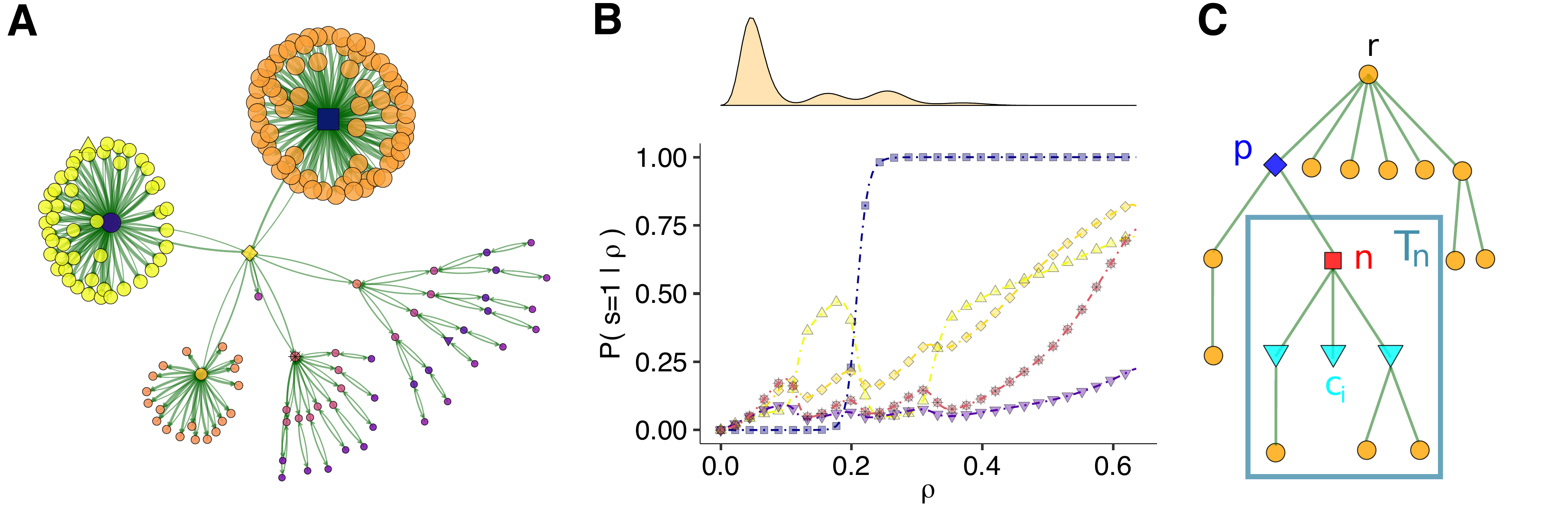}
\end{center}
\caption{A. Exemplary tree with SD parameters. Link strengths are proportional to the ICM weights, node sizes to the activation probabilities. Node colors indicate cluster membership obtained from conditional activation probabilities. B. Conditional activation probability of nodes matching the symbols in A corresponding to the cascade size distribution on top. C. Illustration of variables in SDP.} 
\label{aig:tree}
\end{figure*}

Fig.~\ref{aig:algos} gives an overview of our main contribution: four variants of a message passing algorithm. 
The core is formed by \emph{Subtree Distribution Propagation (SDP)}, which computes the final cascade size distribution based on a rooted tree and ICM as input. 
Every node is visited only once and the algorithm can be parallelized according to the tree structure. 
Starting in the leaves (i.e., nodes with degree $d_n = 1$), each node $n$ sends information about the cascade size distribution of the subtree $T_n$ rooted in $n$ to its parent $p$.
It only requires information by its children as input. 
Finally, the output is constructed in the root. 
The relevant variables are visualized in Fig.~\ref{aig:tree}~C. 

To compute the activation probability of nodes conditional on the final cascade size, \emph{Conditional Subtree Distribution Propagation (conSDP)} adds a backpropagation routine to SDP. 
Information is sent from the root to its children until it reaches the leaves.   
Thus, each node is visited twice in total. 

SDP and conSDP are exact, but are limited to trees. 
To obtain an approximate variant that applies to any (simple) network $G$, we introduce \emph{(conditional) Tree Distribution Approximation (TDA)} as extension of (conditional) SDP. 
The goal is to reduce $G$ to a tree $M$ (for instance a maximum spanning tree (MST)) and run SDP (or conSDP) on $M$.
However, to compensate for the removal of edges and thus dependencies of node activations, we increase the initial activation probabilities $\mathbf{p}$ of the ICM. 
There, we assume that former neighbors $j$ have activated independently  before a node $i$ with a probability $p_{ji}$ that we estimate by Belief Propagation (BP). 
If BP does not converge, we could substitute it by another approach such as the junction tree algorithm.
For simplicity and computational efficiency, we will only consider BP. 
In the following, we detail the information propagation equations of the respective algorithms.%
\subsection{Subtree Distribution Propagation}
\textbf{Intuition}
The derivation of SDP relies solely on combinatorial arguments respecting the right order of activations. 
This order is encoded in messages that a node $n$ sends to its parent $p$ about the number of active nodes $a_n$ in the subtree $T_n$, which is rooted in $n$ (see Fig.~\ref{aig:tree}~C). 
Messages of type $A$ assume that $p$ activates (either before or after the node $n$), while messages of type $B$ consider the case that $p$ does not activate (at least before $n$). 
We need to distinguish these two cases because they enable us to combine the messages received by children efficiently. 

We would like to compute the probability distribution of the number $a_n = \sum_{i \in T_n} s_i$ of active nodes in $T_n$. 
This is given by the sum of $n$'s state $s_n$ and the number of active nodes $a_{c_i}$ in the subtrees that are rooted in $n$'s $k$ children $c_i$: $a_n = s_n + \sum^{k}_{i=1} a_{c_i}$.
The challenge is that the distributions of the summands depend on each other.
Yet, before the activation of $n$, the subtrees $a_{c_i}$ are independent.
The activation of $n$ has an impact on the subtrees but also independently of each other.
Hence, in both cases, the probability distribution of their sum can be computed efficiently as convolution of the right distributions associated with $a_{c_i}$. 
The distributions of $a_{c_i}$ correspond to messages of type $A$ in case that $n$ becomes active and messages of type $B$ in case that $n$ does not.  
In case of the activation of $n$, we have to distinguish two message subtypes, i.e., $A^{\sum}$ and $A^{0}$.
$A^{0}$ is only auxiliary to subtract the probability of an infeasible case, in which $n$ is not activated by any of its children. 

The messages that $n$ receives by its children are then used to compute the distributions of $a_n$ for different cases. 
$p_{na}$ refers to ``no activation'' of $n$, $p_{la}$ to ``late activation'' (or ``no initial activation'') of $n$, and $p_a$ to all cases when $n$ activates.
$p_{la}$ is auxiliary to subtract from $p_a$ the case that no child successfully triggers the activation of $n$. 
At the end of the message passing, the cascade size distribution is computed in the root, where no parent state needs to be considered. 
In case of $a$ final activations, either the root does not activate with $p_{na}(a)$ or activates with $p_{a}(a-1)$ so that $a-1$ other nodes become active.  
The precise algorithm is stated in the following theorem. 
\begin{thm}[SDP]\label{thm:SDP}
The final cascade size distribution $p_C(\rho)$ of an ICM $(\mathbf{p}, \mathbf{W})$ on a tree $G$ with root $r$ and $N$ nodes is given as output of the following message passing algorithm. Starting in the leaves, each node $n$ sends the functions $p_{B_n}(a)$, $p_{A^{0}_n}(a)$, and $p_{A^{\Sigma}_n (a)}$ for $a = 0, ..., N$ as messages to its parent $p$. 
We have
\begin{align*}
& p_{B_n}(0) = 1-p_n,    &&  p_{B_n}(1)   = p_n (1 - w_{np}), \\
& p_{A^{0}_n}(0) = (1-p_n) (1- w_{pn}),    && p_{A^{0}_n}(1)  = (1-p_n)  w_{pn}\\
&  &&     +  p_n (1 - w_{np}),\\
& p_{A^{\Sigma}_n}(0) = (1-p_n) (1- w_{pn}),    && p_{A^{\Sigma}_n}(1)  = p_{A^{0}_n}(1) + p_n w_{np} %(1-p_n)  w_{pn}\\
%& &&  +  p_n. 
\end{align*}
for a leave $n$ (with degree $d_n = 1$). 
Otherwise, define for a node $n$ with $k$ children $c_1, ..., c_k$:
\begin{align*}
& p_{na}(a)  = (1-p_n) p_{B_{c_1}} * \cdots *  p_{B_{c_k}} [a], \\
& p_{la}(a)  = (1-p_n) p_{A^{0}_{c_1}} * \cdots *  p_{A^{0}_{c_k}} [a],\\ 
& p_{a}(a)  =  p_{A^{\Sigma}_{c_1}} * \cdots *  p_{A^{\Sigma}_{c_k}} [a]  -  p_{la}(a),
\end{align*}
where $*$ denotes a convolution. 
An intermediate node $n \neq r$ ($d_n > 1$) with $k_n=d_n-1$ children sends the messages:
\begin{align*}
 & p_{A^{0}_n}(a) =  (1- w_{pn}) p_{na}(a) +   w_{pn} p_{la}(a-1) \\
 & +  (1 - w_{np}) p_{a}(a-1), \  p_{B_n}(a) =  p_{na}(c) +  (1- w_{np}) \\
 & \times p_{a}(a-1), \  p_{A^{\Sigma}_n}(a) =  p_{A^{0}_n}(a)  + w_{np} p_{a}(a-1).
\end{align*}
In the root, $n = r$ with $k_r = d_r$ children, the final cascade size distribution is given as: $p_{C}(a/N) =  p_{na}(a) +  p_{a}(a-1)$.
\end{thm}
Note that all operations can be performed in Fourier space. 
Thus, the convolutions simplify to elementwise multiplication of vectors.
The exact algorithm in Fourier space and a proof are provided in the supplement.

\subsection{Tree Distribution Approximation}
To apply the same principle to a general network $G = (V, E)$ and thus SDP to a spanning tree $M = (V_M, E_M)$ of $G$, we have to regard the direct influence of neighbors $\rm{dn}(i) = \{j \in V \mid  (i,j) \in E,  \; (i,j) \notin E_M\}$ on a node $i$, which are not connected with $i$ in $M$ any more.
For all nodes, those can be estimated by Belief Propagation (BP).

\textbf{Belief Propagation for the ICM}
Given an ICM $(\mathbf{p}, \mathbf{W})$ on a network $G$, the probability $p_{ij} = \mathbb{P}\left(s_i = 1 \parallel s_j = 0 \right)$ that $i$ activates without $j$'s contribution is estimated as
\begin{align*}
& Q_i(t+1)  =  (1 - p_i)\prod_{n \in \rm{nb}(i)} \left(1- w_{ni} p_{ni}(t)\right),  \\ 
& p_{ij}(t+1)  = 1 - \frac{Q_i(t+1)}{1- w_{ji} p_{ji}(t)},  \ \ i, j = 1, ..., N,
\end{align*}
over $t = 0, \ldots, R$ iteration steps.
We initialize with the initial ICM activation probability $p_{ij}(0) = p_i$.
Note that these equations are novel on their own and have been independently derived by Lokhov et al. \cite{lokhov2019scalable}, who prove that they are exact on trees and that the iteration steps $t$ correspond to the time of the cascade process.

Next, $p_{ij} = p_{ij}(R)$ for $j \in \rm{dn}(i)$ are used to adapt the initial activation probabilities $\mathbf{p^{(M)}}$ of an ICM on the spanning tree $M$ (instead of $G$). 
They incorporate the influence of deleted neighbors by assuming that they activate independently before $i$ with $p_{ji}$.

\textbf{ICM on MST} 
We define an ICM $(\mathbf{p^{(M)}}, \mathbf{W^{(M)}})$ on $M$:
\begin{align*}
 & p^{(M)}_i = 1 - (1-p_i) \prod_{j \in \rm{dn}(i)} (1 - w_{ji}p_{ji}), \ \ w^{(M)}_{ij} =  w_{ij} %\ \rm{ for } \  (i,j) \in E_M, %\ \rm{ and } \ w^{(M)}_{ij} = 0  \ \rm{otherwise.}
\end{align*}
for $i = 1, ..., N$ and $(i,j) \in E_M$.

\subsection{Conditional activation probabilities}
The activation probability of a node conditional on the final cascade size is straight forward to compute for the root at the end of SDP (see Thm.~\ref{thm:SDP}) as $\mathbb{P}\left(s_r = 1 \mid C = a/N \right) = p_{a}(a-1)/p_{C}(a/N)$. 
Yet, to obtain the same for every other node $n$ (that we turn into a root), we have to calculate additional messages that parents send back to their children. %to treat them as a root. 
After SDP, only messages from the former parent are missing, so that $p_{B^c_p}(a)$, $p_{A^{0 c}_p}(a)$, and $p_{A^{\Sigma c}_p (a)}$, where $p$ is treated as a child, while $n$ is the new parent. 
Thus, starting in the root $r$, each parent $p$ backpropagates messages to their children $n$, where $\mathbb{P}\left(s_n = 1 \mid C = a/N \right)$ can be computed.

\textbf{BackPropagation for ConSDP (or ConTDA)}
Using the notation of Thm.~\ref{thm:SDP}, each parent $p$ sends to its child $n$:
\begin{align*}
\begin{split}
 %* p^{-1}_{B_{n}}
 & p_{A^{0 c}_p}(a) =  (1- w_{np}) p^{(p)}_{af \setminus n}(a)  +   w_{np} p^{(p)}_{la\setminus n}(a-1) \\
 & +  (1 - w_{pn}) p^{(p)}_{a\setminus n}(a-1), \
 p_{B^c_p}(a) =  p^{(p)}_{af\setminus n} [a] +  (1- w_{pn}) \\
 & \times p^{(p)}_{a\setminus n}(a-1),  \ 
 p_{A^{\Sigma c}_p}(a) =  p_{A^{0 c}_p}(a)  + w_{np} p^{(p)}_{a\setminus n}(a-1), 
 \end{split}
\end{align*}
where the contribution of $n$ in the convolution forming $p^{(p)}_{x}$ is removed (for $x \in \{af, lf, f\}$). 
Note the swap of $n$ and $p$ in comparison with Thm.~\ref{thm:SDP}. 
For all neighbors, children and parent, the messages from SDP are combined with the new one received by the parent as:
\begin{align*}
\begin{split}
& p^{(n)}_{na}(a)  =  p_{na} * p_{B^c_{p}} [a], \  p^{(n)}_{la}(a)  = p_{la} * p_{A^{0 c}_{p}} [a], \\ 
& p^{(n)}_{a}(a)  = (p_{a} + p_{la}) * p_{A^{\Sigma c}_{p}} [a] - p^{(n)}_{la}(a)
\end{split}
\end{align*}
and form the conditional activation probability of $n$ as $\mathbb{P}\left(s_n = 1 \mid C = a/N \right) = p^{(n)}_{a}(a-1)/\left(p^{(n)}_{na}(a) + p^{(n)}_{a}(a-1)\right)$. 
The precise algorithm is stated in the supplementary material.

\begin{figure*}
\includegraphics[width=0.99\textwidth]{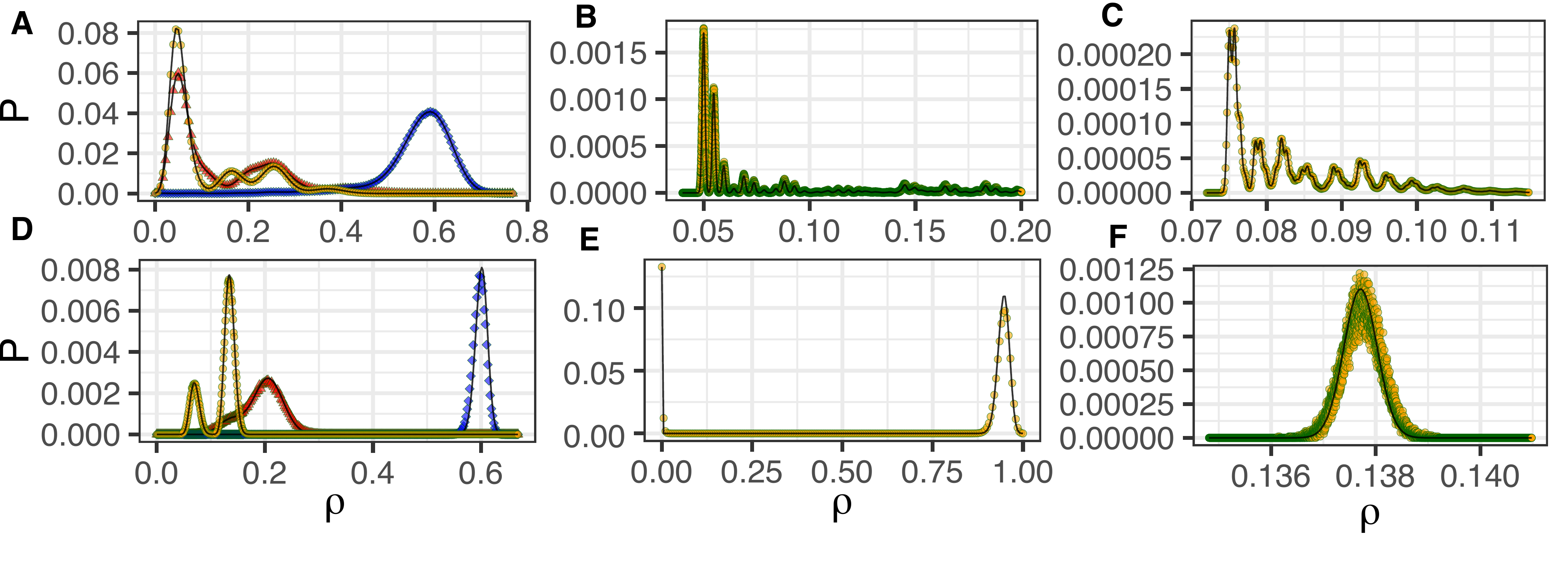}
  \caption{ainal cascade size distribution for TDA (lines) and simulations (symbols). SD (orange circles), DD (blue squares), ED (red triangles). A. Tree ($N=181$). B. Tree ($N=10^5$) with HD. C. Tree ($N=10^6$) with HD. D. Corporate ownership network ($N=4475$). E. Cancer miRNA network with empirical weights ($N = 201$). F. YouTube network ($N=1134890$).}\label{aig:NumTheo}
% \end{wrapfigure}%Youtube network with SD parameters. E.
\end{figure*}
\subsection{Algorithmic complexity}
To validate our claim that our algorithms are usually more efficient than sampling, we have to make sure that we compare with the fastest available sampler.
The precise algorithm is straight forward and stated in the supplement.
Formally, its algorithmic complexity is of the same order as our algorithms for sparse networks, i.e. O($N^2$), but the omitted constant factors differ considerably.

%Both approaches require $O(N^2)$ operations
\textbf{SDP.}
In SDP, each node $n$ is visited once, where mainly the convolution of $d_n-1$ distributions of size $N+1$ in Fourier space has to be computed. 
This can be achieved with $O(d_n N)$ operations.
In total, this adds up to $O(N\sum_n d_n)= O(N^2)$ operations, as $\sum_n d_n = 2*(N-1)$ in a tree. 
Backpropagation requires $O(N^2)$ operations for similar reasons.

\textbf{TDA.}
TDA consists of two additional pre-processing steps: (1) Belief Propagation and (2) the computation of a spanning tree.
(1) BP is linear in the number of edges $O(m)$ and is thus maximally of the same complexity as SDP (because $m \leq N^2$).
(2) A random spanning tree suffices and can be constructed in $O(m)$.
Finding a maximum spanning tree with Kruskal for optimal approximation quality could take longer for dense networks, i.e. $m \log(N)$. 

In practice, for sparse networks and small number of nodes with large degrees, TDA is usually of order $O(N)$, since we can compute the convolutions of distributions with smaller support than $N$ on subtrees. 
%However, also the approximation quality of TDA would be worse in this case. 
Alternatively, one could think of an approximate version of SDP that computes $p_C(\rho)$ for a finite resolution (for instance on an equidistant grid of ${[0,1]}$). 
This could reduce also the worst case complexity to $O(N)$.
With parallelization of computations in nodes that have received the messages by their children, this could be even brought down to $O(h)$, meaning that it is linear in the height of the input tree. 

\begin{figure}[t]
\includegraphics[width=0.5\textwidth]{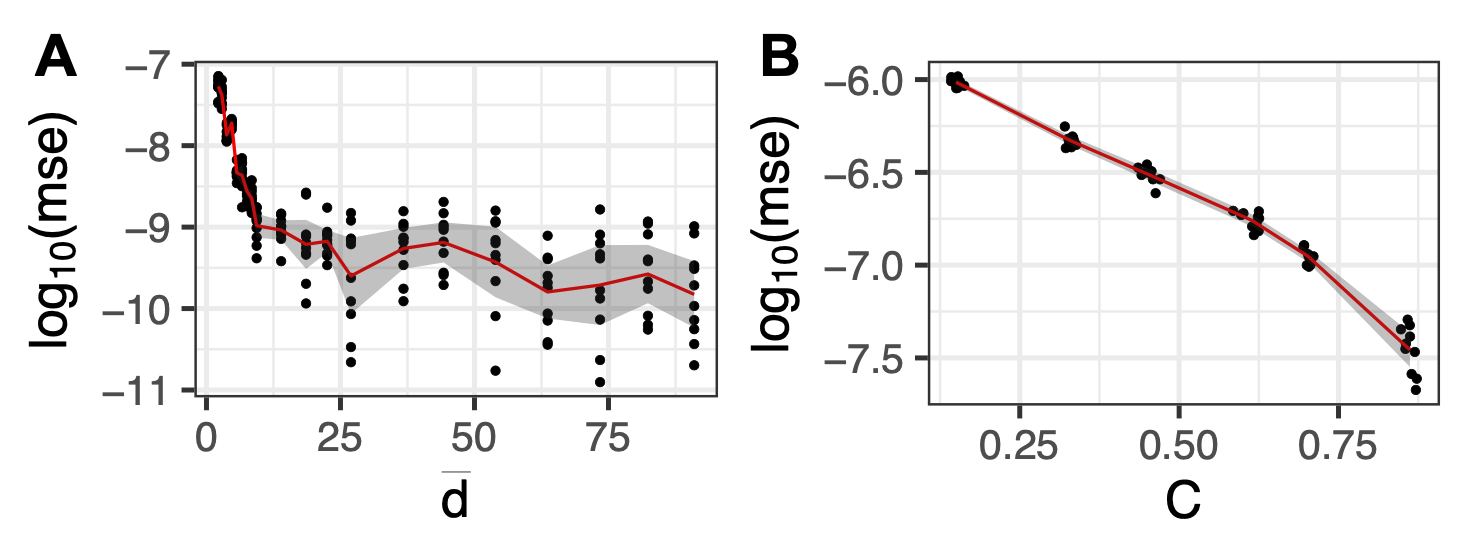}
\caption{Approximation quality of TDA displayed as logarithmic mean squared error (mse) compared with the result of $10^6$ independent Monte Carlo simulations. Each data point corresponds to one experiment. The red line refers to the mean of 10 experiments and the gray area to the associated $0.95$ confidence interval. A. Trend with respect to increasing mean degree $\bar{d}$ of locally tree-like random graphs. B. Trend with respect to increasing clustering coefficient of a random graph.} 
\label{aig:approxQuality}
%\end{wrapfigure}
\end{figure}

\textbf{Sampling.} 
Each sample of a cascade size can be obtained in $O(m)$ operations.
Hence, the total run time is of order $O(N S)$, where $S$ refers to the number of samples.  
How large should $S$ be? 
To approximate the average cascade size, $S=10^4$ is common \cite{Kempe2003} and we compare TDA with this choice in Fig.~\ref{aig:approxQuality}~C.
TDA achieves a speed-up by a factor of $60-500$. 
Yet, depending on the cascade size distribution, $10^4$ samples can lead to poor approximation results.  
The number of required samples is higher, since we have to estimate $N$ free parameters and most of these parameters are small and scale as $1/N$. 
To obtain good enough estimates of each probability, we therefore need at least $O(N)$ samples, where the constant depends on the approximation error. 
This results in a total complexity of $O(mN)$, which becomes $O(N^2)$ for sparse networks. 
According to our experiments, the constant still has to be relatively large. 
\begin{figure*}[t]
\begin{center}
\includegraphics[width=0.85\textwidth]{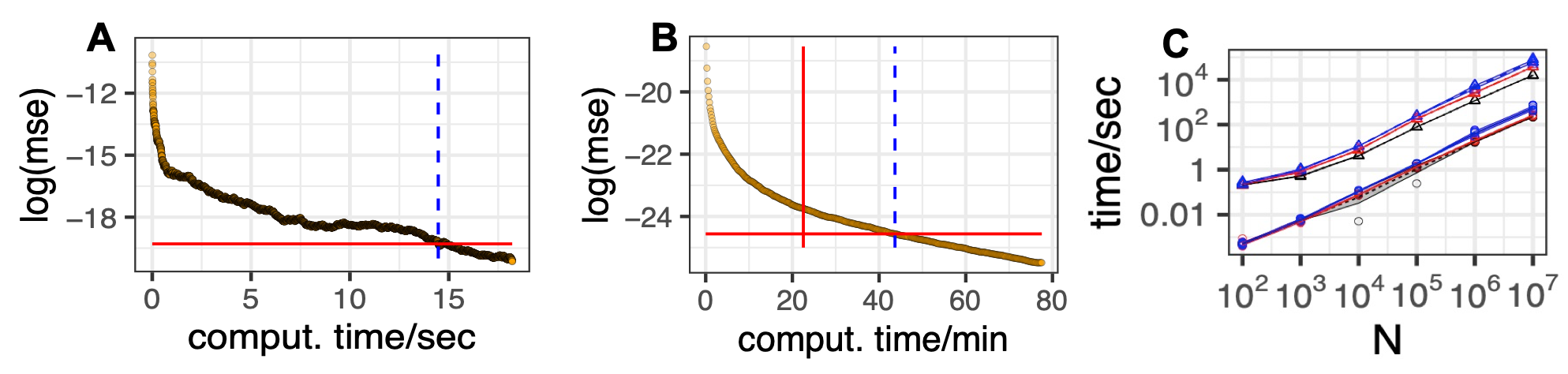}    
\end{center}
\caption{Run time comparisons. A\&B. Mean squared error (mse) of sampling after indicated computation time for SD parameters. The ground truth is defined based on $10^6$ samples. Red horizontal line: mse achieved by TDA. Red vertical line: run time of TDA. Blue dotted line: Time at which sampling achieves the same mse as TDA. A. Tree ($N=181$). TDA takes $0.00078$ seconds. B. YouTube network ($N=1134890$). C. $10$ independent ER-networks of size $N$ are created with average degree $\bar{d}=2.5$ (black), $\bar{d}=4$ (red), $\bar{d}=10$ (blue). Cascade model parameters are drawn iid as $p_n \sim U[1/N, 10/N]$ and $w_{np}\sim U[0.05, 1]$. The run time of sampling the cascade size $10^4$ times (triangles) is compared with TDA (circles) on the same network. Lines refer to averages over $10$ networks and shaded areas to the respective $0.95$ confidence region. On average, TDA is faster by a factor ranging from $68$ to $485$.} 
\label{aig:scaling}
%\end{wrapfigure}
\end{figure*}
\section{Numerical Experiments}
We perform detailed experiments on networks with different properties like size, average degree, and clustering coefficient. 
All computations were conducted on a MacBook Pro with 2.9 GHz Intel Core i9 processor and 32 GB 2400 MHz DDR4 memory. 
Fig.~\ref{aig:NumTheo} provides an overview. 
The first row corresponds to trees of different size, while the second row shows cascade size distributions for real world networks.
The first is a locally tree-like network of corporate ownership relationships \cite{EVAData} with $N = 4475$ nodes, the second a dense correlation network of miRNA expression profiles using data from gastrointestinal cancer \cite{cancer} with $N = 201$ nodes, and the third is a YouTube friendship network \cite{youtube} consisting of $N=1134890$ nodes. 
ICM parameters are varied to demonstrate the diversity of possible cascade size distributions.
Usually, we assume that all nodes activate initially with probability $p_i=0.05$.
For miRNAs, $p_i = 0.01$ to mitigate extensive spreading due to the high network density. 
Unless stated otherwise, we fix the weight parameters to a social dynamics (SD) model with $w_{ij} = 0.05 + 0.5 d_i/d_j/Z$ and $Z = \max_{i,j} (d_i/d_j)$, in which hubs (i.e. nodes with high degrees) are more likely to activate network neighbors but are less likely to become activated by nodes with smaller degrees. 
For variation, we also report results for the following models: 
Financial systemic risk models are often concerned with different risk diversification mechanisms \cite{Battiston10031,Burkholz2015}.
Exposure Diversification (ED) $w_{ij} = 0.05 + 0.5/d_j$ assumes that high degree nodes are difficult to activate by single network neighbors, while in damage diversification (DD) models high degree nodes less likely to infect a network neighbor, for instance, reflected by the choice $w_{ij} = 0.6$ for $d_i \geq d_j$ and $w_{ij} = 0.8$. 
The Hub diversification (HD) model is similar but makes hubs more difficult to activate with $w_{ij} = 0.5/d_j$ for $d_i < 0.1 d_{max}$ and $w_{ij} = 0.4$ otherwise. 
Last but not least, we also consider models, in which the ICM parameters are drawn from random distributions.\\ 
\textbf{Cascade size distributions are relevant.}
Our first observation based on Fig.~\ref{aig:NumTheo} is that the displayed cascade size distributions are often broad and multi-modal.
This can also be the case for large networks.
For instance, Fig.~\ref{aig:NumTheo}~B\&C belong to networks of size $N=10^5$ and $N=10^6$ respectively.
Visually, we also find good agreement between SDP or TDA and sampling for $10^6$ samples. 
%Next, we study the properties of our algorithms in more detail. 
To study this in more detail we first focus on trees, for which our algorithms are provably exact and provide us with ground truth results.
These allow us to estimate the number of required samples for Monte Carlo (MC) simulations to converge and, in comparison, to study the run time and scaling of our proposed algorithms.\\ 
\textbf{How many samples are required?}
The answer must depend on the desired approximation quality and the shape of the cascade size distribution.
Fig.~\ref{aig:scaling}~A shows the evolution of the mean squared error for an increasing number of samples for the small tree with corresponding SD distribution shown in Fig.~\ref{aig:NumTheo}~A. 
TDA and sampling become indistinguishable with $554800$ or more samples.
In comparison, the SD distribution for the YouTube network is more concentrated.
Hence, already $28300$ samples achieve a similar mean squared error as TDA (see Fig.~\ref{aig:scaling}~B).
Yet, TDA is $18790$ times faster in case of the tree, and almost $2$ times faster in case of the YouTube network.\\  

\textbf{Scaling of TDA.}
Without prior knowledge of properties of the cascade size distribution, it is common practice to fix the amount of samples. %(and conduct more simulations later if necessary). 
In this case, sampling has a run-time complexity of $O(N)$, while TDA has formally a complexity of $O(N^2)$. %for a fixed amount of samples. 
Yet, TDA also scales as $O(N)$ for sparse networks with $m = O(N)$ and bounded degrees.   
This is demonstrated in Fig.~\ref{aig:scaling}~C, where we compare the run time for increasing network size $N$ and sampling with $S=10^4$ samples.
$S=10^4$ is a common choice to estimate the average cascade size. 
While this is often not sufficient to estimate the cascade size distribution or the probability of rare events, TDA is still faster by factors ranging from $68-485$.
We also note that TDA scales to large networks consisting of $N=10^7$ nodes and needs less than $5$ minutes to complete in this case.\\
\textbf{Approximation quality of TDA.}
For trees, SDP is exact. 
For locally tree-like structures, i.e. networks with small clustering coefficient and thus negligible number of short loops, we expect that TDA approximates the cascade size distribution well. 
Fig.~\ref{aig:NumTheo} also shows excellent agreement between TDA and sampling with $S=10^6$ for the corporate ownership network and, surprisingly, even for the dense cancer network.
The YouTube network, however, is a difficult case.
BP fails to provide a good estimates of $p_{ij}$ and overestimates the average cascade size.
Hence, we have to define $p_{ij} = p_i$ in TDA and estimate the average cascade size based on a few (i.e. $100$) samples.
All reported results on the YouTube network are for an accordingly shifted cascade size distribution that meets this estimated average.
This combination of TDA and a few sampling steps is still much faster than sampling alone as shown in Fig.~\ref{aig:scaling}~B.
To analyze systematically, how deviations from a locally tree-like network structure influence the approximation quality of TDA, we present Fig.~\ref{aig:approxQuality}.
Specifically, we study the effect of increasing average degree and thus higher network density and increased clustering.
Surprisingly, TDA leads to relatively good approximation results. 
A high average degree is less problematic as long as the network has a small number of triangles or short cycles.
Also a relatively high clustering coefficient does not hamper the approximation results as long as the average degree is small. 
We vary this coefficient in Fig.~\ref{aig:scaling}~B when we generate random graphs consisting of $N=1000$ nodes according to \cite{cluster}, while we try to keep the average degree fixed.  
The exact parameter choices are detailed in the supplementary material.\\ 
\textbf{Conditional activation probabilities.}
The conditional activation probabilities of exemplary nodes of the small tree (Fig.~\ref{aig:tree}A) are shown in Fig.~\ref{aig:tree}B, as obtained by ConSDP.  
Corresponding figures for the other networks can be found in the supplement.
Conditional activation probabilities vary substantially with the cascade size and often increase non-monotonically, different from what might be expected.
They contain rich information not only about the probability of a node to activate but also about its role in the spreading process. 
Big hubs are more predictable and are always active above a certain cascade size. 
Their activation usually marks larger cascades but does not explain the largest.  
These only occur with the activation of nodes that are more difficult to reach by cascades. 
Nodes, which are topological interchangeable, also have an identical conditional activation probability.
The identification of such symmetries is particularly interesting in the analysis of biological networks like the cancer related miRNA network, since these hint towards similar functions of nodes within pathways and thus redundancies in the network. 
In addition, similar conditional activation probabilities translate into similar effects as seeds. 
Therefore, we can interpret the conditional probabilities as node embedding which associates each node $i$ with a vector $v^i$ with components $v^i_{r+1} = \mathbb{P}\left(s_i = 1\mid \rho = r/N \right)$. 
Any clustering algorithm could be employed for a dimensionality reduction based on $v_i$.
For simplicity, we choose kmeans to obtain $15$ clusters.
Node colors in Fig.~\ref{aig:tree} indicate the cluster membership.

\section{Discussion}
The core algorithms that we have derived, Subtree Distribution Propagation (SDP) and conSDP, compute the final cascade size distribution and conditional activation probabilities given a general independent cascade model (ICM) and can therefore replace expensive sampling procedures in related optimization problems. Furthermore, they provide a functional relationship between model parameters and algorithmic output.
On their basis, efficient algorithms that compute gradients can be derived. 
This can enable first order optimization approaches in cases in which only zeroth order optimization is available so far.

In addition to the cascade size distribution, we can compute the activation probabilities of nodes conditional on the final cascade size. 
These are particularly informative in systemic risk analyses, as they allow the focus on extreme events. 
In addition, they provide a node embedding that allows to cluster nodes with similar functionality for the cascade process.

While our algorithms are exact on trees, real world networks usually have additional connections. 
Tree Distribution Propagation (TDA) provides excellent approximation results for locally tree-like networks.

However, if networks consist of multiple short loops, these loops introduce stronger dependencies of activations than TDA can capture. 
Highly connected nodes activate all together with higher probability.
By approximating a denser network with a tree in TDA, we treat some nodes as conditionally independent when they are not and thus underestimate the variance of the cascade size distribution.  
In this sense, we can interpret TDA as variational approach to obtain a proxy for the final cascade size distribution, where TDA captures more dependencies than variational approaches based on BP alone could provide. 

Future work could improve such approximations by combining SDP with sampling.
SDP only requires estimates of cascade size distributions for subgraphs (considering states of parents), which could also be sampled. 
As long as such subgraphs are connected like trees, SDP can be used to combine the distributions. Such an approach could speed up sampling approaches and improve the accuracy of TDA in highly clustered networks. 

Our algorithms could also be extended to allow for distributions on the ICM parameters and could therefore aid robust influence maximization under model parameter uncertainty \cite{robustIM} or Bayesian model learning approaches. 

\section*{Code availability}
The code in R, Python, and C++ is availabe on GitHub:
{https://github.com/rebekka-burkholz/TDA}
\section*{Acknowledgements}
RB and JQ were supported by a grant from the US National Cancer Institute (1R35CA220523).\\ 
We thank Alkis Gotovos for helpful feedback on the manuscript.

\end{document}